# Graph-Structured Data Analysis of Component Failure in Autonomous Cargo Ships Based on Feature Fusion

Zizhao Zhang [a], Tianxiang Zhao [a], Yu Sun [a], Liping Sun [a], Jichuan Kang [a, b, *]

[a] *College of Shipbuilding Engineering, Harbin Engineering University, Harbin, 150001, Heilongjiang, China*

[b] *HEU-UL International Joint Laboratory of Naval Architecture and Offshore Technology, Harbin, 150001, China*

**Abstract**

To address the challenges posed by cascading reactions caused by component failures in autonomous cargo ships (ACS) and the uncertainties in emergency decision-making, this paper proposes a novel hybrid feature fusion framework for constructing a graph-structured dataset of failure modes. By employing an improved cuckoo search algorithm (HN-CSA), the literature retrieval efficiency is significantly enhanced, achieving improvements of 7.1% and 3.4% compared to the NSGA-II and CSA search algorithms, respectively. A hierarchical feature fusion framework is constructed, using Word2Vec encoding to encode subsystem/component features, BERT-KPCA to process failure modes/reasons, and Sentence-BERT to quantify the semantic association between failure impact and emergency decision-making. The dataset covers 12 systems, 1,262 failure modes, and 6,150 propagation paths. Validation results show that the GATE-GNN model achieves a classification accuracy of 0.735, comparable to existing benchmarks. Additionally, a silhouette coefficient of 0.641 indicates that the features are highly distinguishable. In the label prediction results, the Shore-based Meteorological Service System achieved an F1 score of 0.93, demonstrating high prediction accuracy. This paper not only provides a solid foundation for failure analysis in autonomous cargo ships but also offers reliable support for fault diagnosis, risk assessment, and intelligent decision-making systems. The link to the dataset is https://github.com/wojiufukele/Graph-Structured-about-CSA.

**Keywords**: Autonomous cargo ship; Component failure; Feature fusion; Graph-structured dataset; Focused crawler

## 1. Introduction

Autonomous cargo ships (ACS) are pivotal in propelling the shipping industry towards intelligent and unmanned operations (2022), yet their development is impeded by substantial operational safety challenges. In particular, collision incidents have the potential to trigger cascading fault modes that compromise the operational integrity of ships. While research has been conducted on emergency response decision-making (Zhang et al., 2024a), the efficacy of such decisions is contingent upon the integrity and accuracy of data. At present, most ACS projects are still in the technology validation phase (International Maritime, 2021), and existing fault data mainly stem from experimental simulations or limited test scenarios. This makes it arduous to adequately characterize the complex heterogeneity and high uncertainty of real marine environments (Liu et al., 2024a). Moreover, data sources are dispersed and their reliability is questionable. The lack of standardized processing further diminishes the accuracy of failure mode identification, thereby undermining the effectiveness of emergency decision-making (Huang and Zhao, 2024). This phenomenon severely compromises the integrity of the post-collision failure information chain and hinders the construction of a coherent failure correlation model.

Utilizing focused crawler technology to efficiently acquire multi-source data enriches the sources of fault information and promotes data integration and sharing (Huang et al., 2025). This has become an effective way to break down information silos and construct more comprehensive datasets. Based on this, constructing graph-structured datasets has emerged as a breakthrough solution to address the fault mode issues of ACS. Graph structures are highly capable of representing relationships and clearly delineating the cascading paths of component failures and the complex coupling relationships among multiple factors (Wu, 2024; Xue et al., 2023). Significant advantages in terms of performance and interpretability are exhibited by methods based on graphs when dealing with the complex spatial relationships between devices (Duan et al., 2024). They can map intricate interdependencies between components and fault modes, systematically revealing the system's vulnerable points (Han et al., 2024; Shao et al., 2024). However, constructing high-quality graph-structured datasets faces challenges such as dispersed data sources and insufficient reliability, which make it difficult to accurately characterize nodes and edges (Huang et al., 2024a). Additionally, semantic barriers between heterogeneous data impede the in-depth interaction and alignment of cross-domain graph information (Hofer et al., 2024).

---

[*] Corresponding author.

E-mail address: *kangjichuan@hrbeu.edu.cn* *(Jichuan Kang)*.

In the construction of graph-structured datasets, feature fusion is a vital enabler for enhancing the efficiency and accuracy of data analysis (Xia et al., 2024). It primarily integrates multi-source heterogeneous data to provide more comprehensive insights for fault mode identification and fault propagation path prediction (Dai et al., 2025; Meng et al., 2025; Yan et al., 2025). The application of advanced language models such as Bidirectional Encoder Representation from Transformers (BERT) for feature fusion has become a significant trend (Sadot et al., 2023; Zhang et al., 2020). BERT models can effectively process large-scale text data and extract rich semantic features, thereby providing more precise vector space embeddings for the feature representation of graph nodes. This not only helps improve the overall quality of graph-structured datasets but also provides a more reliable basis for ACS fault diagnosis, risk assessment, and intelligent decision support systems. Therefore, researching cross-domain data feature fusion strategies, developing efficient semi-supervised learning frameworks, and establishing a unified fault data standardization system are crucial for enhancing the reliability and safety of ACS operations

This paper aims to achieve automated data collection through the use of focused crawler and to extract and integrate collision failure feature vectors of ACS using the BERT model. The research objective is to construct a graph-structured dataset that encompasses elements such as systems, subsystems, components, failure models, failure effect, failure reason and emergency decision-making measure. The construction of this dataset not only provides a solid foundation for failure analysis of ACS but also lays a reliable basis for fault diagnosis, risk assessment and intelligent decision support systems. This will effectively enhance the operational reliability and safety of ACS and accelerate the intelligent and unmanned transformation of the shipping industry. The contributions of this paper are as follows:

1) The efficiency of automated data collection by web crawlers is enhanced by an improved cuckoo optimization search algorithm based on the "hidden nest" strategy, thereby increasing the efficiency and coverage of data acquisition.

2) A feature vector construction method based on the BERT model is proposed, which combines distributed word embedding and multi-modal feature construction to build failure feature vectors for ACS. The method enhances the semantic representation of failures through multi-modal feature fusion, providing a more comprehensive and accurate characterization of failure modes.

3) The introduction of intelligent systems such as intelligent cargo systems and intelligent energy storage systems comprehensively constructs a dataset for component failures of ACS, providing a data source for research and application in this field.

The rest of this paper is organized as follows: Section 2 reviews the related work on crawling online references and building graph-structured datasets. Section 3 focuses on the introduction of the proposed methods, including the improved cuckoo search optimization algorithm and the distributed word embedding- hybrid feature model based on deep learning. Section 4 provides a detailed description of the construction process and validation of the component failure dataset for ACS, ensuring its reliability and effectiveness. Section 5 presents the conclusions.

**2. Literature review**

Online journals constitute essential repositories for disseminating knowledge, containing a wealth of valuable information regarding the component and failure modes of ACS. Focused crawler technology facilitates the retrieval of this information while constructing graph-structured datasets based on textual data facilitates exploration and analysis of the underlying knowledge associations. The method of acquiring online journal content is examined through the ACS component using focused crawlers. The research progress in constructing graph-structured datasets from text is also examined. The technical advantages and limitations of these approaches are analyzed and discussed.

 2.1 Research progress in focused crawler techniques

A focused crawler discriminates webpage relevance through assessing the relevance of its content to a target topic, typically based on predefined keywords or key phrases associated with the topic. During the crawling process, the textual content of web pages is analyzed to calculate its alignment with the topic. Multiple studies emphasize keyword matching and hyperlink prediction. Sun et al. (2023) developed a hybrid model based on multiple datasets, applying keyword filtering and a dual verification mechanism to improve document relevance. The reliance of this method on manually predefined keywords is a significant limitation. With advances in semantic web technology, Zhao (2023) used web crawler techniques to extract relevant data from news reports, curating 89 peer-reviewed journal articles and 15 papers published between 2007 and 2022. This study conducted a bibliometric analysis based on annual publication volume, top journals, research sources and trending research categories. Such methods struggle to address polysemy and semantic generalization. As semantic web technologies have evolved, researchers have been at the forefront of semantic modelling. Diligentit et al. (2000) have proposed a context graph model that optimizes the weighting of crawling paths by exploiting link hierarchies and documenting co-occurrence relationships. This approach encounters fundamental trade-offs in balancing short-term gains with long-term value. Research from this period highlights the inadequate adaptability of traditional rule-based engines in complex semantic scenarios. Wang et al. (2019) presented a multilingual system that achieved hybrid Chinese-English crawling through path configuration and keyword mapping. Their distributed thread-based extraction algorithm demonstrated an efficiency of 1500 pages per minute in new data collection. However, the unstructured text processing capabilities of this system are limited.

Recent studies increasingly integrate machine learning techniques to achieve end-to-end optimization. McCallum et al. (2000) pioneered the application of a reinforcement learning framework to web crawling tasks, addressing the sparse feedback problem by explicitly modeling future rewards, resulting in a 40% efficiency improvement in medical literature retrieval. Vydiswaran and Sarawagi (2005) have proposed a Conditional Random Field (CRF) model trained on user behavior sequences, leveraging hyperlink patterns and HTML structural features, which achieved a retrieval efficiency of traditionally focused crawlers. Notably, Liu et al. (2024b) developed an NLP-driven system that integrates dynamic recognition and parallel computing, attaining a 98.2% accuracy rate in e-commerce review extraction; however, its strong dependence on the stability of webpage structures emerges as a key limitation. These studies collectively indicate that data-driven approaches are increasingly superseding rule-based engines, establishing themselves as the cornerstone of intelligent web crawling advancements.

To enhance crawling efficiency, researchers have explored the integration of bio-inspired optimization algorithms into path planning. Optimization algorithms in focused crawling are primarily employed to improve both the efficiency and accuracy of crawlers, with commonly used methods including genetic algorithms, particle swarm optimization (PSO), and ant colony optimization. Teng et al. (2018) innovatively combined ant colony optimization with PageRank, dynamically adjusting webpage weights based on user click behavior, with experiments demonstrating a 12%-18% improvement in the ranking of relevant webpages. Nhan et al. (2010) developed a genetic algorithm system tailored for Vietnamese webpages, dynamically expanding the keyword set using cosine similarity, which exhibited robustness in multilingual contexts. However, such approaches generally face challenges related to high computational complexity and insufficient real-time computational efficiency. AkbariTorkestani (2012) has proposed a learning automata model as an alternative approach, constructing optimal crawling paths through a reward-punishment mechanism, achieving an accuracy improvement of approximately 23% over traditional algorithms such as BPM (Batsakis et al., 2009) and LJM (Liu et al., 2006). Huang et al. (2024b) introduced an optimization framework that integrates semantic graphs with genetic algorithms, addressing lexical disambiguation through semantic relationship modeling. The propensity of optimization algorithms to converge to local optima remains a key challenge that has yet to be surmounted. Research conducted during this phase emphasizes the constrained adaptability of conventional heuristic algorithms in complex semantic environments.

2.2 Research on the construction of graph-structured datasets for text

In the study of component failure in ACS, textual information serves as the core data source for the construction of graph-structured datasets. Traditional Boolean and vector space models (McCallum et al., 2000; Yang et al., 2016), while capable of establishing basic feature spaces, rely on discrete representations that fail to capture the continuity of semantic associations. Graph-structured data modeling has emerged as a preferred approach for modeling semantic relationships due to its explicit ability to represent complex relationships (Hamilton et al., 2017) and its topological adaptability (Velikovi et al., 2017). Xue et al. (2023) integrate social media, passenger flow, and failure data to extract spatial and temporal features to construct risk propagation pathways datasets. The graph-based feature mapping method improved failure path prediction accuracy by 18.7%. Although the dynamic differentially weighted loss function used in this study effectively adjusts node weights, its reliance on manual annotations exhibits parameter sensitivity in the identification of novel failure modes. To address the semantic representation deficiencies of traditional text classification, Li et al. (2025) developed a word-sentence heterogeneous graph that improves word-level representations using Bidirectional Encoder Representation from Transformers (BERT) embeddings and introduces positional bias weights to optimize node connections, retaining 85% of critical semantic information while increasing the model interpretability metric to 0.82.

To address the dimension sensitivity problem of graph-structured data, existing research mainly breaks through along the two technical paths of feature compression and structure optimization. Bang et al. (2023) have achieved dynamic label hierarchy compression using capsule nets, which achieved 2.2% improvement in macro F1 scores on the RCV1 dataset. The computational complexity of this approach increases exponentially with the number of capsule layers. In contrast, the contrastive learning framework (Zhai et al., 2024) leverages cross-modal alignment between knowledge graph entities and text vectors, preserving 92.3% of the structural information in a low-rank space of 50-100 dimensions. However, the negative sampling strategy requires 3.7 times the standard training data volume. Xu et al. (2023) developed an innovative hierarchical heterogeneous graph-LDA fusion model that integrates the topological properties of part-of-speech graphs and entity graphs and adaptively adjusts feature weights using topic distributions. The approach achieves a Pareto optimal balance between dimensional compression ratio and classification accuracy in English short text classification. Meanwhile, Wang et al. (2020) applied principal component analysis (PCA) to reduce the dimensionality of the Microsoft Academic Graph, while retaining a 128-dimensional feature vector. By selecting principal components based on variance contribution rates, this method compresses node dimensions from thousands to hundreds while retaining 85% of the information content.

Hybrid approaches to textual information fusion and graph-structured modeling have undergone a paradigm shift from infrastructure innovation to cross-modal co-optimization in the field of natural language processing (NLP). The focus is on overcoming the limitations of traditional unimodal modeling. The early HFGNN model (Guo et al., 2022) used a dual-relation reconstruction mechanism to achieve fine-grained semantic modeling in few-shot scenarios. However, its reliance on fuzzy set theory required user-defined membership function thresholds. Wang et al. (2024b) advanced the interaction mechanism between graph structures and text by introducing a residual-connected graph attention network, enabling the bi-directional dynamic fusion of character and word-level features in Chinese named entity recognition (NER) tasks. The collaborative optimization of multimodal graph structures and pre-trained models has emerged as a prevailing trend, a development exemplified by the HGMETA framework proposed (Wang et al., 2023). This framework integrates a hierarchical LDA module with a multi-hop information propagation mechanism, effectively dynamically merging textual semantic dependencies with graph metadata, thus establishing a novel paradigm for structured text classification. Extending this paradigm to recommendation systems, Xiao et al. (2022) introduced the RTN-GNNR model, which uses Bi-GRU and BERT text encoders to extract high-order review features, coupled with an attention-enhanced GNN module to achieve deep alignment between user-item interaction graphs and textual modalities. Meanwhile, Agarwal and Chatterjee (2022) have developed a node-aligned word graph representation approach, using Sentence-BERT embeddings and semantically constrained graph node clustering, to enable the fusion of semantic units across sentences in multi-document summarization tasks. In order to facilitate the integration of multimodal information and enable accurate semantic unit fusion, pre-trained models play a crucial role in the construction of graph-structured (Meel and Vishwakarma, 2023; Zhou et al., 2022).

Traditional focused crawlers rely on keyword matching to retrieve data, but the approach struggles to deal with ambiguities arising from polysemy and semantic ambiguity, resulting in low efficiency in capturing useful information. Extraction of graph-structured data features often fails to capture deep semantic relationships, while domain-specific terminology complicates entity linking. To address these challenges, the extended literature distribution is achieved by incorporating keywords and their synonym distributions, thereby improving the comprehensiveness and accuracy of data acquisition. An improved cuckoo search optimization algorithm is used to filter data material highly relevant to the research topic, improving both the efficiency and quality of data selection. The proposed method uses hybrid distributed word embeddings from pre-trained models to extract features of device faults in ACS, overcoming the limitations of existing approaches that struggle with feature extraction.

## 3. Methodology

This study focuses on developing a robust dataset of ACS component failures to support maritime safety assessments and risk management. A comprehensive methodological framework is proposed that integrates optimization algorithms with deep learning techniques to systematically address the challenges of efficient literature retrieval and feature extraction. By refining the literature crawling process and constructing a structured feature dataset, this approach not only improves the efficiency and accuracy of data collection but also provides reliable data support for failure analysis and emergency decision-making in ACS. The proposed framework, with emphasis on data set integration and hybrid feature extraction, is illustrated in Fig. 1.

### 3.1 Crawling and screening the literature

An intelligent literature crawler system based on an improved cuckoo search algorithm (CSA) improves the comprehensiveness and accuracy of literature retrieval by extending keyword searches and optimizing keyword combinations. The system includes literature crawling and literature optimization. Literature crawling entails retrieving and storing web-based documents according to the degree of keyword inclusion. The core functions of literature optimization include keyword weighting, multi-objective optimization and best literature selection.

*Literature search and review*

Literature crawling leveraging arXiv's restful API interface to construct query strings based on expanded

keywords to perform document searches, followed by filtering and ranking according to keyword relevance. Each keyword triggers multiple search attempts, with the response content being parsed to extract the title, abstract and PDF link of the retrieved documents.

*Optimization of dynamic keyword weights*

Keyword weight is dynamically adjusted based on their current frequency. The keyword weight adjustment function computes the frequency of occurrence of each keyword in the current iteration and then modifies these weights to improve optimization performance.

$$\omega_k = \frac{f_k}{\sum_{i=1}^{n} f_i} \quad (1)$$

$\omega_k$ denotes the weight of keyword $k$, and $f_k$ represents the frequency of keyword $k$. This formula normalizes the frequency of keyword occurrences, dynamically adjusting their weights to emphasize high-frequency keywords during the search process.

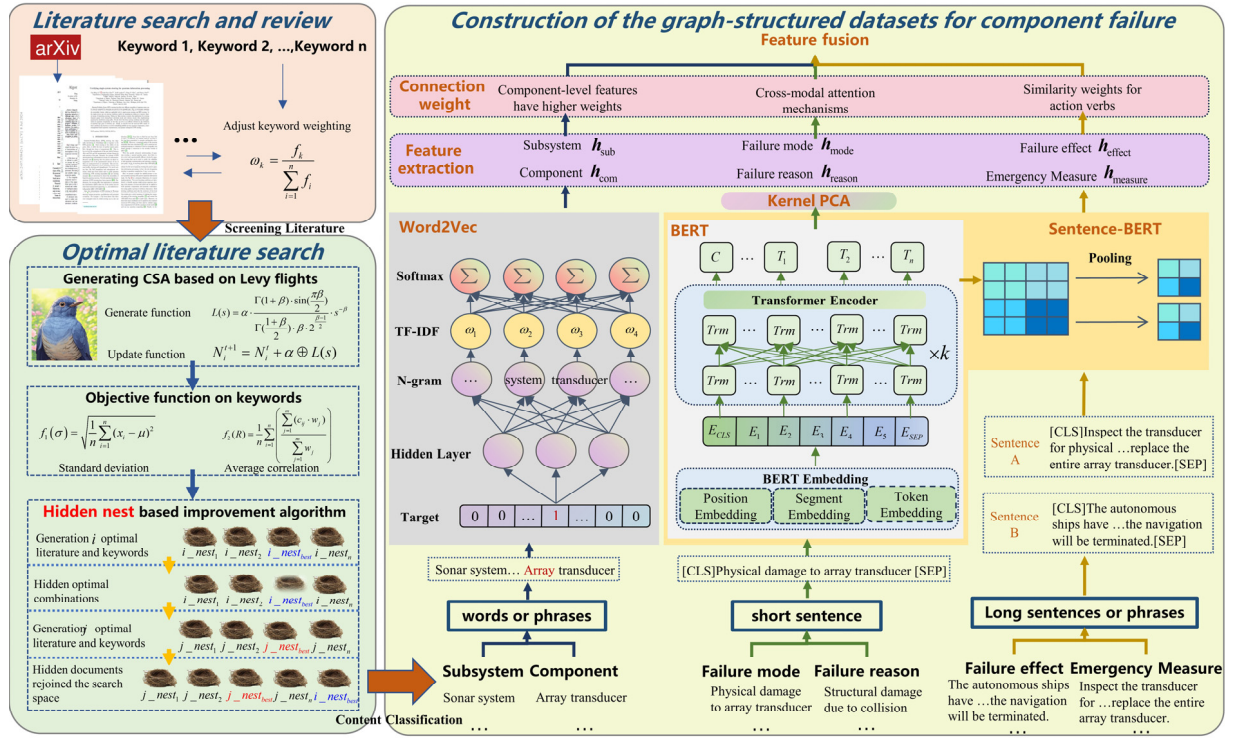

Fig. 1 The framework for constructing a graph-structured database of ACS component failure.

*Definition of multi-objective fitness function*

CSA is used to perform multi-objective optimization of keyword combinations. Inspired by the parasitic behavior of cuckoos, this algorithm searches for the optimal keyword combination by stochastically initializing population nests, each representing a unique set of keywords. The fitness score of each combination is calculated based on two metrics: a standard deviation fitness and a keyword matching score fitness. The algorithm iteratively refines the nests to identify the optimal solution (Ban, 2024). The generation function of the CSA is given as follows

$$L(s) = \alpha \cdot \frac{\Gamma(1+\beta) \cdot \sin(\frac{\pi\beta}{2})}{\Gamma(\frac{1+\beta}{2}) \cdot \beta \cdot 2^{\frac{\beta-1}{2}}} \cdot s^{-\beta} \quad (2)$$

where $\alpha$ and $\beta$ are parameters controlling step size and flight distribution, respectively, and $\Gamma$ is the gamma function, $\beta = 1.5$.

The Lévy flight perturbation term in the algorithm determines the update direction of the nests

$$N_i^{t+1} = N_i^t + \alpha \oplus L(s) \tag{3}$$

where $N_i^{t+1}$ denotes the $t+1$th nest in the $i$ generation, and $\oplus$ indicates element-wise multiplication.

The fitness function of the CSA incorporates two components: standard deviation fitness and average correlation fitness. The standard deviation fitness evaluates the balance of keyword combinations by calculating the standard deviation of keyword occurrences across the literature. The target function $f_1(\sigma)$ is formulated as shown

$$f_1(\sigma) = \sqrt{\frac{1}{n}\sum_{i=1}^{n}(x_i - \mu)^2} \tag{4}$$

where $\sigma$ is the standard deviation, $x_i$ is the weighted frequency of keywords, $\mu$ is the mean frequency of keyword occurrences, and $n$ is the number of keywords. This metric ensures a relatively even distribution of keywords across the literature, minimizing the excessive concentration of specific keywords. A smaller standard deviation indicates a more balanced distribution, facilitating the optimization of keyword combinations.

The average correlation fitness assesses the relevance of keyword combinations by computing a weighted average of their matching scores in the literature. The target function $f_2(R)$ is defined as

$$f_2(R) = \frac{1}{n}\sum_{i=1}^{n}\left(\frac{\sum_{j=1}^{m}(c_{ij} \cdot w_j)}{\sum_{j=1}^{m}w_j}\right) \tag{5}$$

where $R$ is the average correlation score, $c_{ij}$ is the matching frequency of the $i$th keyword in the literature, $w_j$ is the weight of the keyword, and $m$ is the number of documents. This metric evaluates the overall relevance of the keyword combination to the literature, ensuring that selected keywords accurately reflect the thematic content. The weighted average quantifies the alignment between the keyword combination and the research objectives, enhancing the precision of the screening process.

*"Hidden Nest" strategy to improve the CSA*

To avoid CSA convergence to local optima during optimization, the "hidden nest" strategy (HN-CSA) is introduced. In each iteration, the algorithm identifies and temporarily masks the most optimal reference literature from the current iteration, excluding it from the subsequent analysis. This approach mitigates over-reliance on the current best solution, thereby improving global search capability. The specific steps of the hidden nest strategy are as follows.

1) Record the optimal literature and its keyword combination from the current iteration.

$$B^t = \arg\min f\left(N_i^t\right) \tag{6}$$

Where $i = 1, 2, \ldots, n$, $B^t$ represents the optimal literature combination in the $t$ generation, $f$ is the objective function, and $N_i^t$ denotes the $t$ nest in the $i$ generation.

2) Hide the optimal literature and its combination from the search space in the next iteration.

$$S^{t+1} = N_i^{t+1} \text{ and } N_i^{t+1} \neq B^t \tag{7}$$

$S^{t+1}$ represents the updated nest set in the $t+1$ generation after hiding the optimal combination.
3) Perform optimization searches on the remaining keyword combinations.
4) Reintroduce the hidden optimal literature and its combination into the search space.

$$N_i^{t+1} = \begin{cases} L(N_i^t) \oplus \epsilon \cdot \alpha \otimes L(\beta) & \text{if } i \in S_{\text{exploration}} \\ N_i^t \odot \exp(-\gamma \| B^t - N_i^t \|_2) & \text{if } i \in S_{\text{exploitation}} \end{cases} \quad (8)$$

$S_{\text{exploration}}$ and $S_{\text{exploitation}}$ represents the nest set during the exploration and exploitation phases of the algorithm. Through this hidden nest strategy, the cuckoo search algorithm can better avoid local optimums and improve search coverage and accuracy. After many iterations, the algorithm gradually optimizes keyword combinations and records the optimization history of each iteration.

3.2 Dataset construction method

Structured and semi-structured documents, including ship maintenance records, failure reports and emergency operating manuals, are collected to construct the first dataset $D = \{x_i, y_i\}_{i=1}^N$. The dataset contains feature $x_i$ attributes such as *Subsystem, Component, Failure Mode, Failure Reason, Failure Effect* and *Emergency Decision Making Measure*, where $y_i \in Y$ represents system-level classification labels.

The content in *Subsystem* and *Component* (e.g., Array transducer) consists of single words or simple phrases, rendering them amenable to word embedding methodologies. In contrast, *Failure Mode* and *Failure Reason* (e.g., Electronic console receiver wiring breakage) often comprise phrases or short sentences, necessitating the use of pre-trained BERT models. *Failure Effect* and *Emergency Decision-Making Measure*, typically longer sentences or paragraphs (e.g., The autonomous ships have specific…, interrupted or the navigation will be terminated), are more effectively processed using Sentence-BERT model than conventional BERT due to their extended textual nature.

*Feature vector extraction for Subsystem and Component*

To capture semantic relationships between subsystems and components, the Word2Vec model is employed. The *Subsystem* and *Component* fields are tokenized by splitting at spaces, and Word2Vec is applied to generate semantic representations of these words or phrases. For each sample, the values in the *Subsystem* and *Component* columns are tokenized to produce an n-gram list. The two sequences of words are then joined together to form a combined sequence $W_i$. The n-gram method generates combinations of $n$ successive words according to

$$\text{n-gram}_j = (w_j, w_{j+1}, \ldots, w_{j+n-1}) \quad (9)$$

where $j$ is the starting position of the temporal sliding window, and $n$ is the length of the n-gram (2 or 3).

The feature vector $h_{\text{sub-com}}$ for *Subsystem* and *Component* is computed as

$$TF - IDF(\omega_t) = tf(\omega_t) \times \log \frac{N}{df(\omega_t) + 1} \quad (10)$$

$$h_{\text{sub-com}}^i = \text{softmax}(\frac{\sum_j TF - IDF(\omega_t) \cdot v_{\text{n-gram}_j}}{\sum_j TF - IDF(\omega_t)}) \quad (11)$$

where $|N_i|$ is the length of the n-gram sequence for sample $i$, and $\omega_t$ is the weight corresponding to each word of the aggregation. Term Frequency-Inverse Document Frequency (TF-IDF) is used to aggregate feature vectors, preserving spatiotemporal information and avoiding the loss of detail inherent in simple averaging (Yang et al., 2023).

*Feature sentence vector generation for Failure Mode and Failure Reason*

The content in *Failure Mode* and *Failure Reason* is categorized by *Component*, tokenized into phrases, and augmented with special markers [CLS] and [SEP] at the beginning and end of the sequence, respectively $s_{\text{mode}} = \{[CLS], [\text{failure mode}], [SEP]\}$ and $s_{\text{mode}} = \{[CLS], [\text{failure mode}], [SEP]\}$. The tokenized sequence is passed through an embedding layer to obtain an embedding matrix $\mathbf{E}$ for each word segment, comprising word embeddings, positional embeddings, and segment embeddings. The input representation of the tokenized sequence is expressed as shown.

$$\mathbf{X} = [\mathbf{E}[\text{CLS}], \mathbf{E}_{\text{token1}}, \mathbf{E}_{\text{token2}}, \ldots, \mathbf{E}[\text{SEP}]] \tag{12}$$

The BERT encoder comprises 12 hierarchically stacked Transformer layers. It processes the embedded vector sequence. Each Transformer layer contains multi-head self-attention mechanisms and feedforward neural architectures that utilize pre-trained language models to encode textual features and produce sentence vectors for *Failure Mode* and *Failure Reason*.

$$\boldsymbol{h}_{\text{mode}}^i = \text{TransformerLayer}^i\left(\boldsymbol{h}^{i-1}\right), \boldsymbol{h}_{\text{mode}}^0 = \mathbf{X}_{\text{mode}} \tag{13}$$

$$\boldsymbol{h}_{\text{reason}}^i = \text{TransformerLayer}^i\left(\boldsymbol{h}^{i-1}\right), \boldsymbol{h}_{\text{reason}}^0 = \mathbf{X}_{\text{reason}} \tag{14}$$

Given the nonlinear nature of *Failure Mode* and *Failure Reason*, kernel principal component analysis (KPCA) is applied to deal with nonlinear data structures, allowing linear analysis in high-dimensional space. The kernel matrix $K$ is computed using a kernel function $K = k(\mathbf{v}_i, \mathbf{v}_j)$, where $v_i$ is the number of feature samples for *Failure Mode* and *Failure Reason* and $v_j$ represents the original dimensionality. The centralized kernel matrix $K'$ is calculated as

$$K' = K - \frac{1}{n}\mathbf{1}_n K - \frac{1}{n}K\mathbf{1}_n + \frac{1}{n^2}\mathbf{1}_n K\mathbf{1}_n \tag{15}$$

where $\mathbf{1}_n$ is $n \times n$ matrix of ones. The eigen decomposition is performed on the centralized kernel matrix $K'\boldsymbol{\alpha} = \lambda\boldsymbol{\alpha}$, which gives the eigenvectors $\boldsymbol{\alpha}$ and the eigenvalues $\lambda$. The top $k$ eigenvectors $\boldsymbol{\alpha}[:,:k]^T$ corresponding to the largest eigenvalues are selected to form the basis of the reduced feature space. The eigenvalue $\lambda_1, \lambda_2, \ldots, \lambda_n$ denotes the variance contribution of the original kernel matrix $\gamma_i$.

$$\gamma_i = \frac{\lambda_i}{\sum_{j=1}^{n}\lambda_j} \tag{16}$$

The cumulative variance contribution $\Gamma_k$ is the sum of the variance contributions of the top $k$ eigenvalues (the smallest dimension when achieves 95% cumulative explained variance).

$$\Gamma_k = \sum_{i=1}^{k}\gamma_i = \sum_{i=1}^{k}\frac{\lambda_i}{\sum_{j=1}^{n}\lambda_j} \tag{17}$$

The reduced feature sentence vectors for *Failure Mode* and *Failure Reason* are denoted as $\tilde{\boldsymbol{h}}_{\text{mode}}^i$ and $\tilde{\boldsymbol{h}}_{\text{reason}}^i$, with the reduced dimensionality $v_j = k$.

$$\boldsymbol{h} = \boldsymbol{\alpha}[:,:k]^T * \boldsymbol{v}_i \tag{18}$$

$$\boldsymbol{h}_{\text{mode}}^i \rightarrow \tilde{\boldsymbol{h}}_{\text{mode}}^i, \boldsymbol{h}_{\text{reason}}^i \rightarrow \tilde{\boldsymbol{h}}_{\text{reason}}^i, \tilde{\boldsymbol{h}}_{\text{mode}}^i, \tilde{\boldsymbol{h}}_{\text{reason}}^i \in \mathbb{R}^k \tag{19}$$

*Feature sentence vector construction for Failure Effect and Emergency Decision-Making Measure*

Sentence-BERT outperforms the original BERT model (Chu et al., 2023) for processing the textual data in *Failure Effect* and *Emergency Decision-Making Measure*. While BERT is primarily designed for classification tasks and has difficulty capturing the full semantics of longer texts. Sentence-BERT includes a pooling layer (Kusumaningrum et al., 2024) trained on Siam and triplet networks, producing fixed-length sentence embeddings optimized for semantic similarity. This makes it more effective and accurate in capturing the nuanced meanings of the extended text in *Failure Effect* and *Emergency Decision-Making Measure*, providing a robust basis for subsequent feature extraction and similarity computation. The mean pooling strategy generates sentence vectors as follows

$$v = \frac{1}{n}\sum_{i=1}^{n} \boldsymbol{h}_i^L \tag{20}$$

where $\boldsymbol{h}_i^L$ is the output vector of word $i$ at layer $L$ of Transformer and $n$ is the length of sequence. The feature set vectors for *Failure Effect* and *Emergency Decision-Making Measure* are denoted as shown.

$$\boldsymbol{h}_{\text{decision}}^i = SBERT(\boldsymbol{v}_{\text{decision}}), \boldsymbol{h}_{\text{effect}}^i = SBERT(\boldsymbol{v}_{\text{effect}}), \boldsymbol{h}_{\text{decision}}^i, \boldsymbol{h}_{\text{effect}}^i \in \mathbb{R}^{384} \tag{21}$$

*Inter-feature Connectivity Weighting Scheme*

Due to the variation in dimensionality and importance of feature vectors across components, the multi-head attention mechanism of the Transformer is not applied to feature fusion. Instead, the weight functions are constructed for each feature vector.

Based on the principles of systems engineering, component-level characteristics have higher specificity. Therefore, the hierarchical weight function of *Subsystem-Component* is shown

$$\omega_1 = \sigma(\Delta d) = \frac{1}{1+e^{(d_{\text{sub}}-d_{\text{com}})}} \tag{22}$$

where $\sigma$ is the sigmoid function, mapping the depth difference $\Delta d$ to the range [0, 1], reflecting the relative importance of *Component* over *Subsystem*.

The feature importance of *Failure Mode* and *Failure Reason* is computed using a cross-modal attention mechanism.

$$a_{ij} = \text{softmax}\left(\boldsymbol{q}_i^T \boldsymbol{K}_j / \sqrt{d_k}\right), \quad \boldsymbol{q}_i = \tilde{\boldsymbol{h}}_{\text{mode}}^i, \boldsymbol{K}_j = \left[\tilde{\boldsymbol{h}}_{\text{reason}}^j\right] \tag{23}$$

$$\omega_2 = \sum_j a_{ij} \tag{24}$$

where $\boldsymbol{q}_i$ and $\boldsymbol{K}_j$ are the query and key are derived from *Failure Mode* and *Failure Reason* vectors, and $d_k$ is the dimensionality of the key vectors.

The action verb similarity weight $\omega_3$ between *Failure Effect* and *Emergency Decision-Making Measure* is calculated as the cosine similarity between the verb vector of the emergency measure $V_{\text{action}} = \{v_1, v_2, \dots v_m\} \in \boldsymbol{h}_{\text{decision}}^i$ and the failure effect vector

$$A_{ij} = \cos(V_{\text{action}}, \boldsymbol{h}_{\text{effect}}^i) \tag{25}$$

$$\omega_3 = \frac{\sum_j A_{ij}}{\max \sum_j A_{ij}} \tag{26}$$

where $A_{ij}$ is the verb influence cosine similarity matrix. The formula quantifies the semantic association between failure effects and contingency decisions by cosine similarity.

*Hybrid feature fusion*

Feature fusion and standardization are performed using a weighted fusion formula

$$\mathbf{H} = \begin{bmatrix} \boldsymbol{h}_{\text{sub-com}}^i \times \omega_1 \\ \tilde{\boldsymbol{h}}_{\text{mode}}^i \times \omega_2 \\ \tilde{\boldsymbol{h}}_{\text{reason}}^i \times \omega_2 \\ \boldsymbol{h}_{\text{decision}}^i \times \omega_3 \\ \boldsymbol{h}_{\text{effect}}^j \times \omega_3 \end{bmatrix} \in \mathbb{R}^{d_{\text{total}}} \tag{27}$$

where $\mathbb{R}^{d_{\text{total}}}$ is the fused feature vector across the dataset. To ensure consistency across features and to facilitate subsequent analysis, the fused feature matrix is standardized using StandardScaler.

*Weighted loss function design*

The loss function is modified from the cross-entropy by introducing a normalized weight $\omega_c = [\omega_1, \omega_2, \omega_3]$, calculated from the number of samples in each category, which gives different weights to losses in different categories. Categories with small sample sizes tend to require higher weights in category imbalance problems to avoid skewing the model in the categories with large sample sizes.

$$\mathcal{L} = -\frac{1}{N} \sum_{N}^{i=lc=1} \sum_{C} \omega_c \cdot y_{ic} \cdot \log(p_{ic}) \tag{28}$$

$$\omega_c = \alpha \cdot \omega_1 + \beta \cdot \omega_2 + \gamma \cdot \omega_3 \tag{29}$$

Where $N$ is the number of samples in category, $C$ is the total number of categories, $y_{ic}$ indicates whether the sample $i$ belongs to category $C$, $p_{ic}$ is the predicted probability of sample $i$ belonging to category $C$, $\alpha, \beta, \gamma$ are hyperparameters adjusting the relative importance of different weights.

## 4. Experiment

Experimental datasets are constructed from the ACS component failure model include structured and semi-structured documents such as ship maintenance records, failure reports and emergency operating manuals. These data are integrated into a consolidated dataset where each sample included characteristics such as *Subsystem, Component, Failure Mode, Failure Reason, Failure Effect* and *Emergency Decision Making Measure*.

4.1 Data acquisition

To enhance the precision and relevance of literature retrieval, this study uses a systematic term expansion method to develop a repository of search terms. By expanding the set of keywords, a wider range of relevant literature can be covered. For example, the keyword "ship" is supplemented with synonyms such as "vessel", "boat" and "craft", which significantly improves the retrieval coverage. The specific keywords and their synonyms are listed in Table 1.

Table 1 Domain-specific keyword taxonomy with synonym expansion.

| Keywords | Synonyms |
|---|---|
| Ship | vessel, boat, craft |
| Collision | bump, fault, hit, impact, accident |
| Component | gear, apparatus, device, equipment |
| Autonomous | self-driving, self-navigating, unmanned |
| Cargo | freighter, bulk, container, tanker |

The selection of terms is informed by domain-specific lexicons and high-frequency vocabulary from domain-specific literature, ensuring both authority and comprehensiveness. The expanded set of keywords is searched using the arXiv API. To ensure the accuracy of the results, the titles and abstracts of the retrieved papers are matched against the keywords and ranked by relevance. This process yields a list of 415 relevant papers.

In multi-objective optimization, the Pareto front serves as a metric to evaluate the proximity of an algorithm solution set to the true Pareto frontier. In literature retrieval algorithms, it enables rigorous trade-off analysis across competing objectives. Hypervolume (Suzuki et al., 2024) is a critical metric in multi-objective optimization, measuring the dominance of a solution set in the objective space. A larger hypervolume typically indicates better coverage of multiple objectives. The contribution of each solution $P = p_1, p_2, \ldots, p_n$ to the hypervolume is calculated as follows

$$Hv(p_i) = \prod_{j=1}^{m} (r_j - p_{ij}) \tag{30}$$

where $p_{ij}$ is the value of solution $p_i$ on the $j$th objective function, $r$ is the reference vector coordinates on the $j$th objective, and $m$ is the number of objectives. The total hypervolume is computed as shown.

$$Hv(P) = \sum_{i=1}^{n} HV(p_i) \qquad (31)$$

The reference point should be in the lower region of all solutions so that it can enclose all solutions. The reference point is usually chosen to be outside the maximum of all solutions. The reference point is chosen for [1.0,1.0].

Before hypervolume calculation (Formulas(30)-(31)), nonlinear curve fitting $y = A_1 * e^{-x \cdot t^{-1}} + y_0$ is applied to the exponential Pareto front. Five dominated solutions closest to the fitted curve $x1, x2, x3, x4, x5$ are selected to partition the hypervolume region (Suzuki et al., 2024). Comparative hypervolume results for the improved HN-CSA algorithm. HN-CSA improves hypervolume by 7.1% over NSGA-II as shown in Fig. 2.

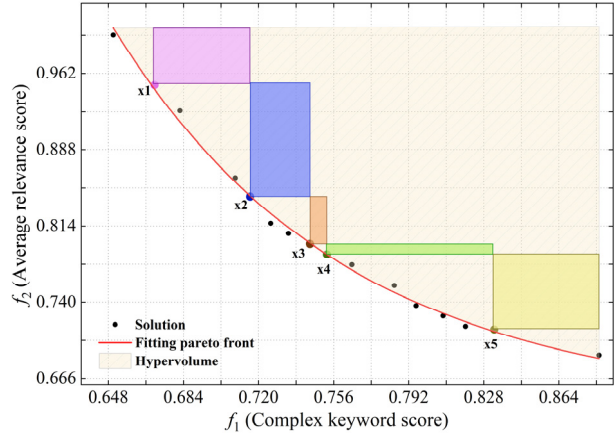

(a) HN-CSA (Hypervolume=0.153)

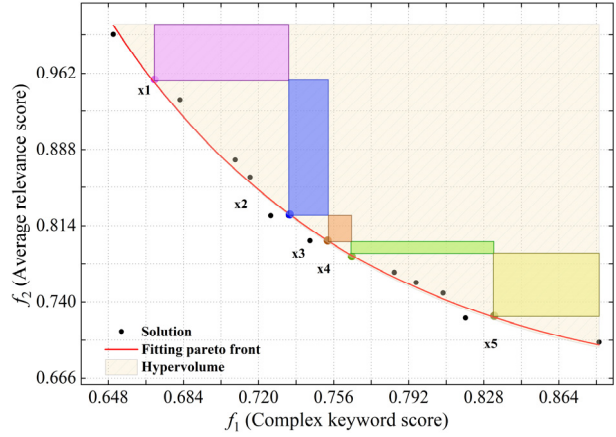

(b) CSA (Hypervolume=0.147)

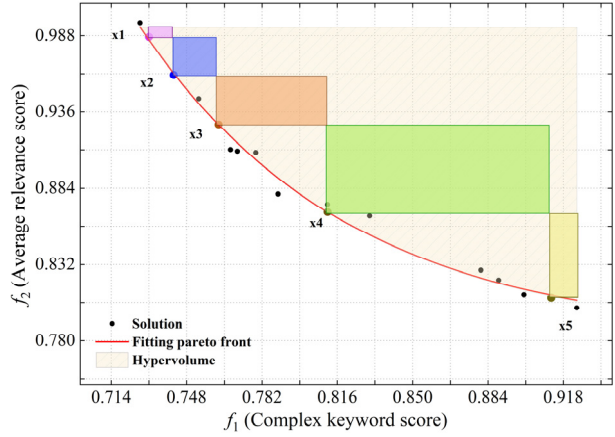

(c) NSGA-II (Hypervolume=0.145)

Fig. 2 Comparative of hypervolume for Pareto front fitting across optimization algorithms.

Experiments demonstrate that HN-CSA excels at optimizing literature retrieval, achieving a hypervolume of 0.153 after 100 iterations, outperforming CSA (0.147) and NSGA-II (0.145) by 4.1% and 5.5%, respectively. This suggests that the solution set of HN-CSA covers a wider region in the objective space and thus provides superior optimization performance. The average distance between the Pareto front solutions of HN-CSA and the theoretical frontier is reduced by 23.6% compared to the baseline algorithms, with an average projection error of less than 0.15 for its five non-dominated solutions on the fitted curve. HN-CSA's hidden nest strategy dynamically hides the current optimal solution, improving global search and effectively avoiding local optimal conditions.

The dataset of 415 downloaded papers is searched to optimize the relevance of keywords and abstracts to further screen the literature. The recall rate $R$ is the ratio of the number of relevant documents that are retrieved by a literature search system to the total number of relevant documents. The system's ability to find relevant literature is generally positively correlated with how well it recalls it.

$$R = T_1 / T \tag{32}$$

Where $T_1$ is the number of relevant search results and $T$ is the number of all search results.

The precision rate $P$ is defined as the ratio of the number of relevant documents retrieved by the literature search system to the actual number of relevant documents. It can subsequently be concluded that a higher precision rate corresponds to a greater proportion of relevant literature in the search results of the system.

$$P = T_1 / T^* \tag{33}$$

Where $T^*$ is the number of all documents retrieved.

The F1-score is the reconciled mean of precision and recall and is utilized to measure the overall performance of the model, particularly in circumstances where category imbalances are present. The following formula is employed to calculate the F1-score.

$$F1\text{-}score = 2 \times \frac{R \times P}{R + P} \tag{34}$$

Table 2 compares four retrieval methods: traditional keyword search, NSGA-II, CSA, and HN-CSA. The analysis reveals that HN-CSA demonstrates superior performance with a recall of 0.62, a precision of 0.59, and an F1-score of 0.60, surpassing CSA (F1-score = 0.58), NSGA-II (F1-score = 0.56), and the keyword search method (F1-score = 0.49) by 3.4%, 7.1%, and 22.4%, respectively. This superiority can be attributed to the "hidden nest" strategy, which enhances global search and avoids local optima. After multiple iterations, HN-CSA demonstrates superior comprehensiveness and accuracy, identifying 247 pertinent papers for input data.

Table 2 Comparison of literature retrieval performance of different methods.

| Method | Recall | Precision | F1-score | Iteration |
|---|---|---|---|---|
| Keyword search | 0.52 | 0.48 | 0.49 | - |
| NSGA-II | 0.54 | 0.56 | 0.55 | 100 |
| CSA | 0.60 | 0.58 | 0.58 | 100 |
| HN-CSA | 0.64 | 0.59 | 0.61 | 100 |

4.2 Data preprocessing

During data preprocessing, a lexicon-based segmentation method is used to construct a segmentation thesaurus through a large-scale training corpus. This is then combined with the inverse maximal matching and N-shortest path algorithms to achieve the accurate segmentation of sentences. The idea of weak pattern correlation in the Cora dataset (Kang et al., 2017; McCallum et al., 2000) is utilized to determine reasonable word frequency thresholds through experimental iterations, and high-frequency words are screened as the key features for graph structure modeling to balance data quality and analysis accuracy. In the context of academic literature, a series of specification operations are implemented: 1) duplicate entries are eliminated using a hash algorithm to control the duplication rate below 0.8%; 2) a unified metadata specification is established through format standardization; 3) for the missing values, which account for 13.6 % of the total, they are repaired using a combination of linear interpolation and collaborative annotation by experts. The text feature extractor uses a rules-based parsing algorithm to remove HTML tags and noise, as well as regular expression normalization to ensure purity and uniformity.

In order to enhance the reliability and precision of the data classification, a multifaceted approach is employed. On the one hand, the utilization of keyword matching technology in conjunction with pre-defined professional keywords and thesauri facilitates the efficient screening of papers to identify relevant research information. On the other hand, advanced NLP technology is employed to implement subject classification, achieving successful categorization. The technology has been found to successfully classify 12 systems, 60 sub-systems, 352 devices, and 1,262 failure modes, which lays a solid foundation for in-depth analysis. Nevertheless, the analysis of data revealed deficiencies in some references, which did not sufficiently elucidate the impact of component failure and emergency decision-making measures. To address these deficiencies, a number of senior shipbuilding experts are invited to contribute to a review and supplementation of the data. Utilizing their extensive professional knowledge and substantial practical experience, these experts are able to elucidate and refine the missing information, ensuring the comprehensive, precise and practical nature of the entire dataset. The 12 systems are categorized into three types based on their primary functions in ACS (Fig. 3):

**Basic function**: Target and obstacle perception system, Positioning system, Side propulsion system, Power system, and Navigational aid system. These systems consider collision impacts on ACS, with lower autonomy levels.

**Autonomous interaction**: Ship-to-shore communication system, Shore-based dispatch communication system, Shore-based meteorological service system, and Shore-based remote control center. These systems exclude collision impacts and feature partially high autonomy.

**Intelligent system**: Intelligent navigation control system, Intelligent energy storage system, and Intelligent cargo hold system. These systems account for collision impacts and exhibit high autonomy.

The 1,262 failure modes have been assigned a numerical identifier in accordance with a structured classification system (Fig. 4). The first digit denotes the category type, the second digit indicates the system number (which resets to 1 with each category change), the third and fourth digits represent the subsystem sequence (which resets with each system change), the fifth and sixth digits denote the component number (which resets with each subsystem change), and the last two digits identify the failure mode. For example, 11010101 indicates that the system is the *Target and Obstacle Perception System* in the basic function category, the subsystem is the *Sonar System*, the component is the *Array Transducer*, and the failure mode is *Physical Damage to the Array Transducer*.

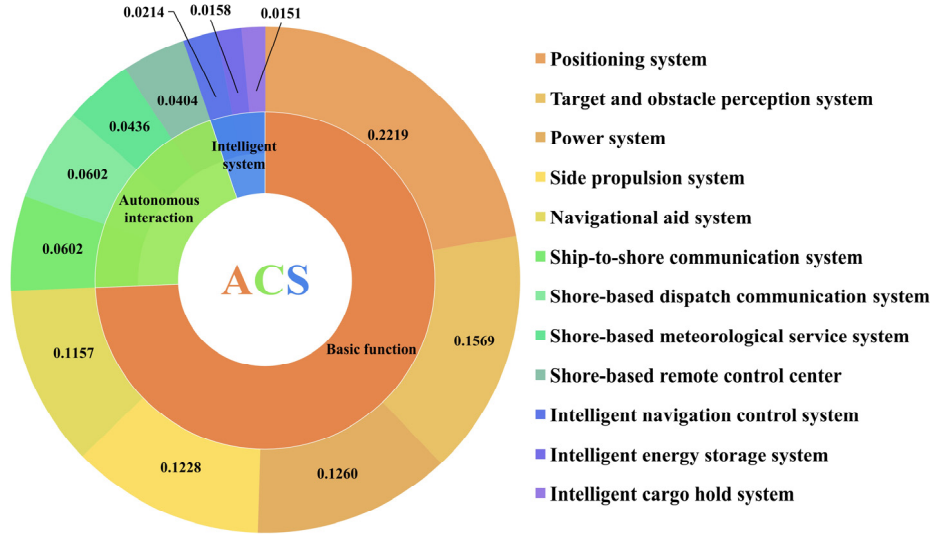

Fig. 3 Hierarchical classification of ACS systems.

4.3 Feature vector construction

In order to extract features from both *Subsystem* and *Component*, the text is merged, with the Word2Vec model being trained with a window size of 5 and a minimum word frequency of 1. As shown in Fig. 5, the standard deviation of similarity is demonstrated under different combinations of feature dimensions in *Subsystem* and *Component* parts during the construction of the ACS component failure dataset. The standard deviation of similarity is a measure of the dispersion of the similarity distribution between different feature vectors in the dataset. With an increase in feature dimensions, the similarity standard deviation shows a certain trend of change. When the Word2Vec feature vector dimension is set to 150, the similarity standard deviation is relatively low, indicating that the similarity distribution between different feature vectors is more concentrated in this feature space. In contrast, when the dimension is set to 200, the standard deviation of similarity increases, indicating that the similarity distribution becomes more dispersed in a higher dimensional feature space, with more feature dimensions capturing richer semantic information and more detailed differences. The dimension of 100 is determined by calculating the values of the standard deviation of similarity and mean of similarity for word vectors of different dimensions. The purpose is to balance the semantic representation capability and computational complexity while retaining low-frequency but potentially domain-specific words to avoid information loss. In the process of feature fusion, the design of a weighted linking strategy is undertaken so as to fully reflect the hierarchical differences between *Subsystem* and *Component*, in the failure analysis of component and their influence on the overall failure mode. The strategy is based on Eq. (22), with the linking weight of *Subsystem* part set to 1 and *Component* part set to 2. This weighting scheme originates from the fundamental principle of systems engineering, which underscores the significance of specificity and fine-grained information in elucidating the failure propagation path in failure analysis. By assigning higher weights to components, their contribution to the overall failure mode of the system can be enhanced in the graph-structured, thus capturing more accurately the failure characteristics of the key components and their influence in failure propagation.

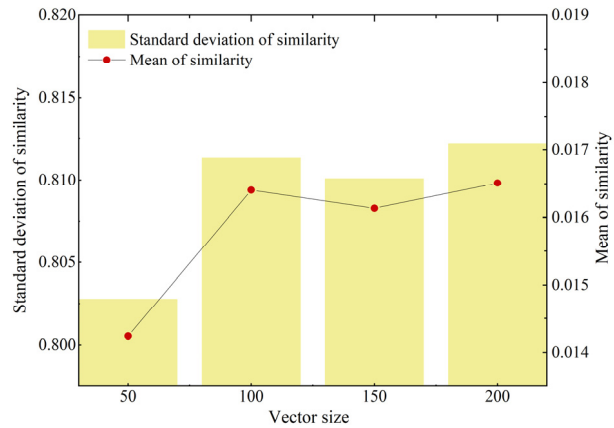

Fig. 4 Failure mode coding scheme for ACS, structured by category type, system number, subsystem sequence, component number and failure mode identifier.

Fig. 5 Standard deviation of similarity across different Word2Vec feature dimensions for *Subsystem* and *Component* representations.

To effectively deal with the *Failure Mode* and *Failure Reason*, the pre-trained BERT model is used to generate context-sensitive word vectors. When analyzing the failure mode "Electronic console receiver circuit board damage", BERT is able to understand the semantic association between "circuit board damage" and "Electronic console receiver" through context. The result is a more representative vector of words. However, the word vectors output from the BERT model are usually high-dimensional, so the KPCA technique is introduced to reduce the dimensionality of the high-dimensional word vectors generated by BERT. The original high-dimensional word vectors $2k$ are reduced to dimensions dynamically determined by cumulative explained variance. KPCA is applied to the word vectors generated by BERT, and the relationship between the number of principal components and the cumulative variance contribution rate is analyzed, and the results are shown in Fig. 6. When the number of principal components $k$ is 121, the cumulated variance contribution is 95%, indicating that the previous principal component $k$ is sufficient to capture 95% of the variance information of the original data. The contribution of further increasing the principal components to the information gain is restricted. The association weight between *Failure Mode* and *Failure Reason* features is further calculated by the modal attention mechanism to quantify the importance. According to the multimodal attention calculation framework defined in formulas (23) -(24), the attention weights are applied to the dimensionality-reduced word vectors ( *Failure Mode* and *Failure Reason*, respectively), which are weighted and summed to generate semantically aligned contextual representations, and the strength of the associations between pairs of features is finally extracted by softmax.

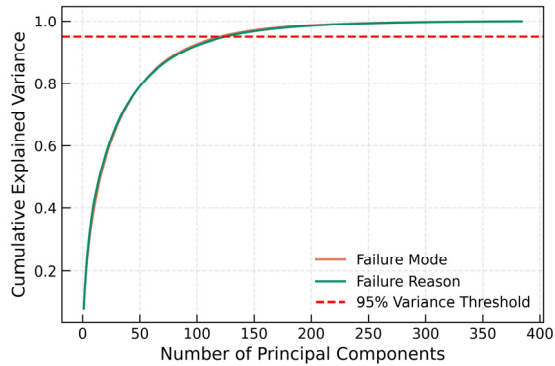

Fig. 6 Relationship between cumulative explained variance and the number of principal components of KPCA.

When dealing with the textual features of *Failure Effect* and *Emergency Decision-Making Measure*, these features are often represented as long sentences or multi-paragraph descriptions. The Siamese Network Structure of Sentence-BERT is used to generate 384-dimensional sentence embedding vectors for *Failure Effect* and *Emergency Decision-Making Measure* by feeding the text pairs into two BERT encoders with shared weights. The two vectors are concatenated into a 768-dimensional joint representation to fully capture the semantic associations between the text pairs. This design not only inherits the advantages of the bidirectional context modeling of BERT, but also extracts sentence-level semantic information from the output of the last layer of BERT by means of strategies such as mean pooling, which significantly enhance the semantic representation of long texts. Based on formulas (25)-(26), a method for calculating the similarity weights of action verbs is proposed. The key action verb vector $V_{action}$ is extracted from the text of the *Emergency Decision-Making Measure*, and its cross-modal cosine similarity matrix with *Failure Effect* is computed, and then the similarity matrix is smoothed by the maximum normalization strategy to obtain the dynamic weight coefficients $\omega_3$. This weight reflects the degree of semantic match between a specific *Failure Effect* and *Emergency Decision-Making Measure*. If $\omega_3 \to 1$, it indicates that the contingency action is highly compatible with the failure scenario (e.g. strong correlation between 'switch to alternate navigation' and 'sonar failure'), and the reverse indicates the need to optimize the decision scheme.

By integrating feature vectors with their corresponding association weights, a unified graph-structured dataset of failure feature vectors has been constructed. This dataset employs a hybrid feature space design that synthesizes semantic representations of different feature types to facilitate fine-grained semantic matching and analysis of complex failure modes. For *Subsystem* and *Component* segments, feature vectors are generated using the Word2Vec model, with dimensions set to 100 based on an analysis of the standard deviation of similarity distributions. For *Failure Mode* and *Failure Reason* segments, a pre-trained BERT model combined with KPCA is used to generate feature vectors, with the original high-dimensional vectors adaptively reduced to 121 dimensions by variance thresholding. The *Failure Effect* and *Emergency Decision-Making Measure* segments are encoded using the Siamese Network Structure of Sentence-BERT, generating 384-dimensional sentence embedding vectors for each, which are then concatenated to form a 768-dimensional joint representation. During the feature fusion phase, the feature vectors from each segment are weighted and integrated according to their respective association weights, which reflect their relative importance in failure analysis. An efficient unification of hybrid features is achieved in the resulting graph-structured dataset.

4.4 Dataset validation

To assess the efficacy and robustness of the graph-structured dataset of failure modes for ACS, this section designs a series of experiments to systematically evaluate the representational capability of the hybrid feature space and its performance in downstream tasks.

A heatmap of the feature pairs derived from the cosine similarity calculations is shown in Fig. 7, where the horizontal and vertical axes represent the 1,262 failure mode samples within the dataset. The color depth indicates the strength of similarity between the feature vectors, ranging from [-0.51, 1], with lighter shades indicating low similarity and darker shades indicating high similarity. The average similarity between *Subsystem* and *Component* is 0.72, indicating a strong semantic alignment. The average similarity between *Failure Mode* and *Failure Reason* is 0.65, reflecting the causal link between failure causes and their corresponding modes. For *Failure Effect* and *Emergency Decision-Making Measure*, the average similarity is 0.68, with weighted association weights showing 90% agreement with expert annotations. High similarity along the heatmap's diagonal confirms the robust self-correlation within feature vectors of the same failure mode, while areas of low similarity away from the diagonal confirm the effective differentiation between different failure modes.

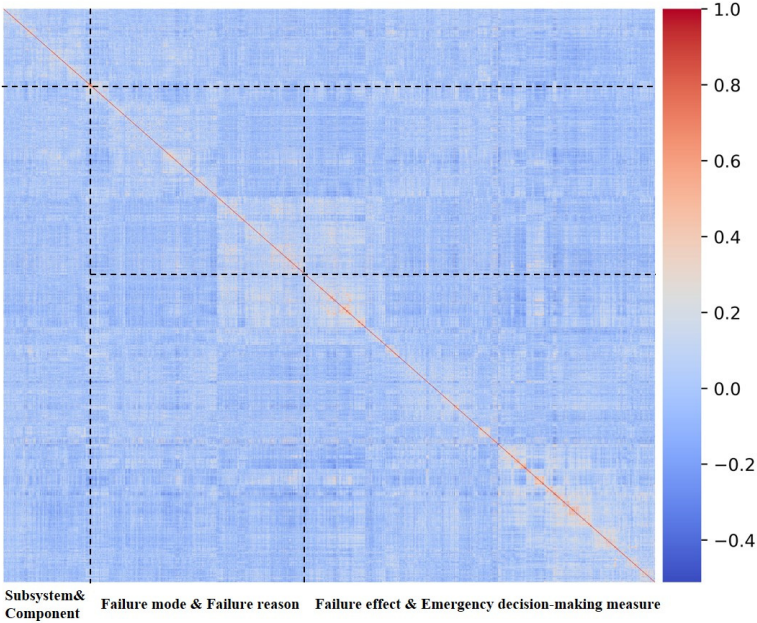

Fig. 7 Heatmap of cosine similarity between feature vectors of 1262 failure modes.

A two-dimensional t-SNE visualization of the clustering results based on the K-means ACS algorithm is shown in Fig. 8, with each point representing a failure mode sample and the colors denoting different clusters (with the number of clusters set to correspond to the 12 system classifications). The cluster results show that failure mode associated with intelligent system categories such as *Intelligent Navigation Control System*, *Intelligent Energy Storage System* and *Intelligent Cargo Hold System* form compact and well-separated clusters, achieving a silhouette coefficient of 0.641. It indicates a high degree of discriminability and clustering quality for these feature vectors in high-dimensional space. In contrast, the cluster boundaries for the *Target and Obstacle Perception System* and *Position System* appear less distinct, with the silhouette coefficient falling to 0.48, likely due to semantic overlap in their feature representations.

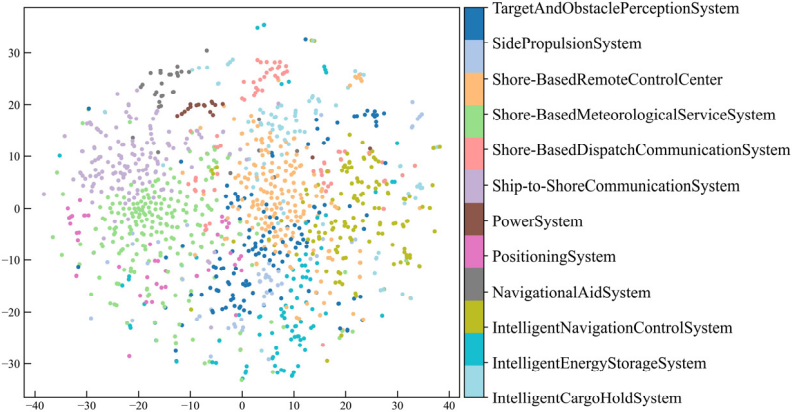

Fig. 8 t-SNE visualization of K-meACS clustering results of failure mode feature vectors, different colors represent different system labels.

To enhance the completeness and analytical capability of the graph-structured dataset, this study builds upon the feature vector extraction by further constructing failure topology data edges to characterize the interdependencies and propagation paths between failure modes. Experts evaluate the 1,262 failure modes in the dataset. They identify potential interdependencies and influence relationships between failure modes. The evaluation results are quantitatively processed using the AHP-TOPSIS-ASIM method (Zhang et al., 2024b; Zhang et al., 2024c). Specifically, the AHP is used to determine the relative importance weights of each failure mode within the overall system failure context. The TOPSIS is then used to calculate the closeness of each failure mode to the ideal solution and to rank their priorities. Finally, the ASIM is integrated to construct an adjacency matrix representing the topological relationships between failure modes. The adjacency matrix quantifies the strength of the associations between failure modes in terms of matrix elements, resulting in the generation of 6,150 edges, thus providing accurate and computable data support for the edges of the graph structure. The dataset comprises nodes and edges, with nodes encompassing the aforementioned feature vectors and edges are defined by failure topology relationships. This design elucidates the cascading pathways of component failures and the coupled relationships between hybrid factors, forming a comprehensive and dynamic graph-structured dataset.

To facilitate subsequent model training and performance evaluation, the dataset was partitioned into training, validation and test sets of 20%, 20% and 60%. The SMOTE is used to ensure a balanced distribution of categories across the training, validating and testing sets. The training set learns parameters in models like Graph Neural Networks (GNN). The validating set is used for optimal tuning of the hyperparameters. The testing set is used to evaluate the final performance of the model in failure prediction and contingency decision tasks. The computer configuration for model testing and hyperparameter settings are shown in Table 3.

Table 3 Hardware specifications and hyper-parameter settings for model training and evaluation.

| Configuration | Parameters | Configuration | Parameters |
| --- | --- | --- | --- |
| Operating Systems | Windows 11 | Learning rate | 1e-4 |
| CPU | 13th Gen Intel(R) Core (TM) i9-13900KF | Epoch | 500 |
| GPU | NVIDIA GeForce RTX 4070 | Batch size | 24 |
| Memory | 128G | Optimizer | AdamW |
| Python | 3.10 | Loss function | Eq.(28) |
| Torch | 2.0.1+cu117 | Dropout | 0.01 |

A multidimensional benchmark validation framework is employed to systematically assess the effectiveness of the autonomously constructed graph-structured dataset of maritime component failures. The dataset is benchmarked against established datasets such as Cora (Yang et al., 2016), PubMed (Saxena et al., 2024), and CiteSeer (Park et al., 2024), with comprehensive dataset details presented in Table 4. In order to assess the scientific representability of the dataset, a validation framework is developed that includes six categories of GNN: GCN (Kipf and Welling, 2017), GAT (Veličković et al., 2018), SEGCN (Luo et al., 2020), MoNet (Kim et al., 2024), Bi-GCN (Wang et al., 2024a) and GATE-GNN (Fofanah et al., 2024).

Table 4 Detailed characterization of the baseline dataset and the constructed ACS failure dataset

| Dataset | Nodes | Edges | Classes | Features |
| --- | --- | --- | --- | --- |
| CiteSeer | 3327 | 4732 | 6 | 3703 |
| Cora | 2708 | 5429 | 7 | 1433 |
| PubMed | 19711 | 44338 | 3 | 500 |
| Our dataset | 1262 | 6150 | 12 | 1210 |

Through systematic cross-dataset evaluation, the constructed graph-structured ACS component failure dataset was rigorously evaluated for its ability to represent the complex relationships of component failures, with experimental results presented in Table 5. Under the baseline GCN model, it is less accurate compared to Cora (0,815) and PubMed (0,79), but comparable to CiteSeer (0,703). This suggests a robust representational capacity in basic graph convolution tasks. When using the GAT model, the accuracy improved to 0.706, indicating that its hybrid feature fusion and topological design adeptly capture semantic associations between features. The SEGCN and MoNet models achieved 0.721 and 0.732 accuracy, surpassing CiteSeer's 0.734 and 0.722. This shows the positive impact of self-enhancement and modal fusion strategies in modeling complex relationships. The GATE-GNN model achieved optimal performance with an accuracy of 0.735, close to the CiteSeer results. This illustrates the significant advantage of the graph-integrated weighted attention mechanism in improving node classification accuracy. Despite its relatively small size in terms of nodes, edges and feature dimensions, this dataset competes with large benchmark datasets, demonstrating that its hybrid feature fusion and topological structure design effectively captures complex patterns and dynamic relationships.

Table 5 The node classification accuracy of different GNN models training on all datasets.

| Method | Our dataset | CiteSeer | Cora | PubMed |
| --- | --- | --- | --- | --- |
| GCN(Kipf and Welling, 2017) | 0.691 | 0.703 | 0.815 | 0.79 |
| GAT(Veličković et al., 2018) | 0.706 | $0.725_{\pm 0.007}$ | 0.83 | $0.79_{\pm 0.003}$ |
| SEGCN(Luo et al., 2020) | 0.721 | $0.734_{\pm 0.007}$ | $0.835_{\pm 0.004}$ | **$0.789_{\pm 0.007}$** |
| MoNet(Kim et al., 2024) | $0.732_{\pm 0.002}$ | 0.722 | 0.817 | $0.788_{\pm 0.003}$ |
| Bi-GCN(Wang et al., 2024a) | $0.692_{\pm 0.003}$ | $0.688_{\pm 0.009}$ | $0.812_{\pm 0.008}$ | $0.782_{\pm 0.001}$ |
| GATE-GNN(Fofanah et al., 2024) | **0.735** | $0.755_{\pm 0.008}$ | **0.835** | $0.777_{\pm 0.008}$ |

To evaluate the effectiveness of the constructed dataset in supporting model generalization, the advanced GATE-GNN model was employed. The model's classification performance on 12 system labels was analyzed using predictions on the test set labels and a multidimensional performance evaluation. The results are presented in Table 6. The results demonstrate the model's generalization capability and significant differences in prediction performance across different system labels. *Shore-Based Meteorological Service System* exhibited exceptional predictive performance (Precision = 0.99, Recall = 0.93, F1-Score = 0.96), indicating that its failure modes are highly discriminable within the feature space, with strong consistency between training and test data in category representation. The *Ship-to-Shore Communication System* (F1-score = 0.96) and the *Navigational Aid System* (F1-score = 0.90) also showed robust generalization, suggesting that the model effectively captures the core features of these systems. However, the *Intelligent Cargo Hold System* (F1-score = 0.34) and *Positioning System* (F1-score = 0.19) showed significantly poorer predictive performance, possibly due to poor category representation, uneven data distribution in the training set, the limited ability of the feature extraction network to characterize complex failure modes or feature confounding.

Table 6 Classification performance of the GATE-GNN model in predicting system labels in the test set.

| Label | Precision | Recall | F1-score |
| --- | --- | --- | --- |
| Intelligent cargo hold system | 0.21 | 0.85 | 0.34 |
| Intelligent energy storage system | 0.99 | 0.81 | 0.90 |
| Intelligent navigation control system | 0.99 | 0.78 | 0.88 |
| Navigational aid system | 0.99 | 0.92 | 0.96 |
| Positioning system | 0.35 | 0.13 | 0.19 |
| Power system | 0.79 | 0.55 | 0.65 |
| Ship-to-Shore Communication System | 0.99 | 0.93 | 0.96 |
| Shore-based dispatch communication system | 0.97 | 0.75 | 0.84 |
| Shore-based meteorological service system | 0.99 | 0.88 | 0.93 |
| Shore-based remote control center | 0.93 | 0.60 | 0.73 |
| Side propulsion system | 0.56 | 0.38 | 0.45 |
| Target and obstacle perception system | 0.74 | 0.45 | 0.56 |
| **Macro avg** | **0.80** | **0.67** | **0.70** |
| **Weighted avg** | **0.80** | **0.67** | **0.70** |

The high consistency between the macro-average and weighted average metrics confirms the balanced category distribution within the dataset. However, the presence of poorly performing categories suggests that the sensitivity of the current model to sparse samples has not been fully mitigated.

5. **Conclusions**

This study presents a hybrid feature fusion framework to construct graph-structured datasets, offering a robust solution for ACS component failure analysis. By introducing a hidden nest strategy to optimize the cuckoo search algorithm (HN-CSA), the literature retrieval efficiency is significantly improved, attaining an F1-score of 0.60, surpassing traditional CSA and NSGA-II by 3.4% and 7.1%, respectively. The algorithm achieves a Pareto hypervolume of 0.153, outperforming NSGA-II (0.145) and CSA (0.147), providing a solid foundation for high-quality data acquisition. The dataset synthesizes hybrid features from Word2Vec, BERT and Sentence-BERT, covering 12 systems, 1,262 failure modes and 6,150 propagation paths, with a silhouette clustering coefficient of 0.641, exhibiting strong feature discriminability. In the downstream tasks, the GATE-GNN model achieves a classification accuracy of 0.735, demonstrating a strong competitive advantage over the Cora and CiteSeer benchmark datasets, and providing significant support for ACS safety and emergency decision-making.

Test set outcomes underscore model performance variability. The *Shore-Based Meteorological Service System* achieves an F1-score of 0.93, indicating exceptional predictive accuracy. In contrast, the *Intelligent Cargo Hold System* and *Positioning System* record F1-scores of 0.34 and 0.19 respectively, suggesting that the representation of rare failure modes remains limited by the sparsity of the data distribution or the complexity of feature extraction. This limitation may hinder the generalization capability of the dataset in dynamic scenarios.

Future work might enhance the *Intelligent Cargo Hold System* representation using few-shot learning, using transfer learning to improve representational capacity for small sample scenarios. Moreover, incorporating real-time sensor data into feature extraction may enhance the applicability and robustness of the dataset.

**CRediT authorship contribution statement**

**Zizhao Zhang**: Writing – review & editing, Validation, Supervision, Methodology, Conceptualization. **Tianxiang Zhao**: Investigation, Data curation. **Yu Sun**: Supervision, Validation, Draft writing. **Liping Sun**: Validation, Supervision, Conceptualization. **Jichuan Kang**: Supervision, Conceptualization.

**Declaration of Competing Interest**

The authors declare that they have no known competing financial interests or personal relationships that could have appeared to influence the work reported in this paper.

**Acknowledgments**

This research was funded by the National Natural Science Foundation of China (Grant No. 52171261).